\documentclass{article}

\usepackage[final,nonatbib]{neurips_2023}

\usepackage[utf8]{inputenc} 
\usepackage[T1]{fontenc}    
\usepackage{hyperref}       
\usepackage{url}            
\usepackage{booktabs}       
\usepackage{amsfonts}       
\usepackage{nicefrac}       
\usepackage{microtype}      
\usepackage{xcolor}         
\usepackage{wrapfig,lipsum,booktabs}
\usepackage{graphicx}
\usepackage{enumitem}
\usepackage{caption}
\usepackage{subcaption}

\usepackage{amsmath}
\DeclareMathOperator*{\argmin}{arg\,min}

\title{Improving Domain Generalization in Contrastive Learning using Adaptive Temperature Control}

\author{%
  \textbf{Robert Lewis}$^{*}$ \\
  Massachusetts Institute of Technology\\
  Cambridge, MA, USA \\
  \texttt{roblewis@mit.edu} \\
  \and
  \textbf{Katie Matton}$^{*}$ \\
  Massachusetts Institute of Technology\\
  Cambridge, MA, USA \\
  \texttt{kmatton@mit.edu} \\
  \and
  \textbf{Rosalind W. Picard} \\
  Massachusetts Institute of Technology\\
  Cambridge, MA, USA \\
  \and
  \textbf{John Guttag} \\
  Massachusetts Institute of Technology\\
  Cambridge, MA, USA \\
  {\small $^{*}$Equal contribution.}
}

\begin{document}

\maketitle

\begin{abstract}
Self-supervised pre-training with contrastive learning is a powerful method for learning from sparsely labeled data. However, performance can drop considerably when there is a shift in the distribution of data from training to test time. We study this phenomenon in a setting in which the training data come from multiple domains, and the test data come from a domain not seen at training that is subject to significant covariate shift. We present a new method for contrastive learning that incorporates domain labels to increase the domain invariance of learned representations, leading to improved out-of-distribution generalization. Our method adjusts the temperature parameter in the \textit{InfoNCE} loss -- which controls the relative weighting of negative pairs -- using the probability that a negative sample comes from the same domain as the anchor. This upweights pairs from more similar domains, encouraging the model to discriminate samples based on \textit{domain-invariant} attributes. Through experiments on a variant of the MNIST dataset, we demonstrate that our method yields better out-of-distribution performance than domain generalization baselines. Furthermore, our method maintains strong in-distribution task performance, substantially outperforming baselines on this measure.
\end{abstract}

\section{Introduction}

Unsupervised contrastive learning is an effective strategy for learning meaningful representations from unlabeled data. It typically uses a form of self-supervision where augmented versions of the same data instance (i.e., \textit{positive pairs}) are encouraged to be close in representation space while different instances (i.e., \textit{negative pairs}) are pushed apart. This encodes the idea that if instances differ only in the augmentation applied, then they should be mapped to the same representation. Hence, contrastive learning produces representations with invariance to the chosen augmentations. 

This is useful when we know \textit{a priori} the types of perturbations to which the representations should be invariant. When this holds, pre-training a feature extraction network (i.e., encoder) using contrastive learning helps to build models that are generalizable. However, in many real-world settings, it is not feasible to specify augmentations that account for the distribution shifts that occur at test time. For example, it is unclear what augmentations simulate the types of shifts that occur when using a medical imaging model on data from a new hospital. When the data augmentations used \textit{do not} include the shifts that occur at test time, the performance of contrastive pre-training drops significantly. 

We study this problem in the setting in which the training data (i.e., data for pre-training and supervised training) come from multiple domains and the goal is to generalize to an \textit{unseen} domain. Deviating from most work in domain generalization, we assume that most of the training data from each domain is \textit{unlabeled}, which elevates the importance of finding a robust pre-training strategy. This problem setting is important because labeled data can be expensive to collect.

We propose a new method for contrastive learning that leverages domain labels to produce representations with increased invariance to domain differences. Our work builds on two key findings. First, when computing the contrastive loss, it can be beneficial to upweight the contributions of negative samples that come from the same domain as the \textit{anchor} instance \cite{zhang2022towards,cheng2020subject}. Doing so encourages the model to use \textit{domain-invariant} information to discriminate instances. Second, the temperature parameter $\tau$ in the \textit{InfoNCE} loss (cf. Eq.~\ref{eqn:info-nce}) controls the penalty given to different negative samples; a smaller $\tau$ results in a larger penalty for \textit{hard negatives} (i.e., negatives that are more similar to the anchor instance) \cite{wang2021understanding}. Accordingly, we propose to use the temperature parameter to upweight samples that appear to come from similar domains. We introduce an adaptive, domain-aware temperature parameter whose value we adjust based on the probability that a negative sample comes from the same domain as the anchor. We estimate this probability using a domain discriminator trained on top of the \textit{current} representation space, allowing for sample weights to change as the amount of domain-discriminating information in the representation space changes. We discuss related work in Appendix~\ref{sec:app_related_work}. 

We evaluate our method on a variant of the MNIST dataset that we designed to enable careful control over different aspects of the data generating process. This allows us to empirically examine how different factors (e.g., types and strengths of domain shifts) impact the representations learned and to build intuition to guide future theoretical analysis. We find that under most experimental settings, our method yields better out-of-distribution (OOD) and in-distribution (ID) performance when compared to a set of baseline methods that use alternative strategies for encouraging domain invariance.

\section{Improving Robustness to Covariate Shift in Contrastive Learning using Adaptive Temperature Control}\label{sec:method}

\textbf{Domain Generalization from Sparsely Labeled Data.} Let $X \in \mathcal{X}$ be the features and $Y \in \mathcal{Y}$ be the labels. We consider data from multiple domains, where each domain $d$ is characterized by a joint distribution $P^d(X,Y)$ and a marginal distribution $P^d_X(X)$. We assume access to data from training domains $d \in \{1, \ldots, d_{\text{tr}}\}$, which are each associated with two datasets: 
\begin{itemize} 
    \item[(1)] \textit{unlabeled}: $D^d_{UL} = \{(x_i^d, d): x_i^d \overset{\mathrm{iid}}{\sim} P^d_X(X) \}_{i=1}^{n^d_{UL}} $;
    \item[(2)] \textit{labeled}: $D^d_{L} = \{(x_i^d, y_i^d, d): (x_i^d, y_i^d) \overset{\mathrm{iid}}{\sim} P^d(X,Y)\}_{i=1}^{n^d_{L}}$,
\end{itemize} 
where (1) is available during pre-training and (2) is used \textit{downstream} for supervised learning. We assume the data are sparsely labeled; i.e., $n^d_{UL} \gg n^d_L$. We aim to learn a model $\theta^*$ that performs well on an $\textit{unseen}$ domain $d_{\text{te}}$; i.e., for a given loss function $\ell(X,Y;\theta)$ we want to find $\theta^* = \argmin_\theta \mathbb{E}_{(X,Y) \sim P^{d_{\text{te}}}}[\ell(X,Y;\theta)]$.

To expect that a model could perform well on an unseen domain, we need to make an assumption about invariances that hold across domains. We focus on the \textit{covariate shift} setting: the marginal distributions of $X$ differ across domains, but the conditional distributions of their labels $Y$ given $X$ are the same. This is stated formally as: 
\begin{align}
\begin{split}
    P^d_X(X) &\neq P^{d'}_X(X)\quad \wedge\\
    P^d_{Y|X}(Y|X) &= P^{d'}_{Y|X}(Y|X), \qquad \forall d,d' : d \neq d'
\end{split}
\end{align}

\textbf{Contrastive Pre-training.}
We focus on the popular \textit{SimCLR} framework \cite{chen2020simple}, but note that the modifications we propose also apply to other contrastive learning formulations that use InfoNCE loss variants e.g., MoCo and SupCon \cite{he2020momentum,khosla2020supervised}. Formally, let $f: \mathbb{R}^p \rightarrow \mathbb{R}^k$ be an encoder network that maps from inputs $x$ to low-dimensional embeddings, with $k \ll p$. Let $A$ be a set of data augmentations. To train $f$, a random mini-batch of N examples is sampled. For each sample, $x_i$, two augmented versions are generated using randomly sampled data augmentations: $a(x_i)$ and $ a'(x_i)$ for $a, a' \sim A$. Let $z_i = f(a(x_i); \theta_f), z_i' = f(a'(x_i); \theta_f)$ be the outputs of the encoder on augmented versions of $x_i$. A separate loss term for each embedding $z_i$ is computed, where it is treated as the $anchor$ and positive and negative pairs are defined with respect to it. The standard, unadjusted \textit{InfoNCE} loss \cite{oord2018representation} for anchor $z_i$ is defined as:
\begin{equation}\label{eqn:info-nce}
    \ell_{i} = -log \frac{exp({sim(z_i, z'_i)/\tau})}{\sum_{j=1}^Nexp(sim(z_i, z'_j)/\tau)}
\end{equation}
where $sim(z_i, z'_j)$ is the \textit{cosine similarity} between embeddings and $\tau$ is the temperature parameter. The temperature controls the relative weighting of negative samples in a batch; a smaller $\tau$ results in a higher penalty on \textit{hard negatives} -- i.e., those that are more similar to the anchor instance \cite{wang2021understanding}.

\textbf{Pre-training Objective with Adaptive Temperature.} We propose a new method for contrastive pre-training that adjusts the temperature, $\tau$, to increase the penalty associated with negative samples that appear to come from a similar domain to that of the anchor. This leads to domain invariance by encouraging the model to learn to discriminate samples based on domain-invariant information. 

We want the relative weightings of samples to be adaptive to the current embedding space. This is important because it means that as the embedding space is refined, we can leverage the training signal from a larger number of negative samples in the batch to improve the model. To achieve this, we train a domain discriminator network $g: \mathbb{R}^k \rightarrow \mathbb{R}^{|D|}$ that takes as input the embeddings $z_i$ produced by the contrastive encoder. This allows us to compute the probability that an embedding $z_i$ belongs to domain $d$ as: $P(D = d | z_i) = softmax(g(z_i);\theta_g)$. We optimize the domain discriminator using cross entropy loss. We do \textit{not} backpropagate the loss from the domain discriminator to the contrastive encoder, so there is no competition with the contrastive loss on the encoder weights (more details on training the domain discriminator are in Appendix~\ref{sec:app_train_domain_discriminator}). 

We use the domain discriminator to compute the probability that anchor embedding $z_i$ and negative sample $z_j$ come from the same domain; we denote this $w_{ij}$. This can be done in multiple ways, resulting in two variants of our method:
\begin{itemize} 
    \item \textbf{Domain-Weighted Negatives:} we treat the domain of the anchor example $z_i$ as known and use a soft label for $z_j$ only. We compute
    $w_{ij} = P(D_j=d_i|z_j) = P(D|z_j)[d_i]$.
    \item \textbf{Domain-Weighted Pairs:} we use a soft label for the domains of both the anchor and its negative. We compute
    $w_{ij} = P(D_i = D_j | z_i, z_j) = \sum_d P(D_i = d | z_i) P(D_j = d | z_j) $.
\end{itemize} 

We then use these weights, $w_{ij}$, to compute an adaptive temperature for each negative pair:
\begin{equation}\label{eqn:domain-adjust-info-nce}
    \tau_{ij} = max(\tau_\alpha + \tau_\beta(\frac{1}{N_D} - w_{ij}), \tau_{min})
\end{equation}
Where $\tau_\alpha$ is a baseline temperature, $\tau_\beta$ controls the degree to which we adjust the temperature based on the domain probabilities, and $\tau_{min}$ bounds the temperature to not go below a certain value to prevent numerical instabilities. $N_D$ is the number of domains and $\frac{1}{N_D}$ is the uniform probability when the domain discriminator is maximally uncertain about the domain a sample belongs to. We apply this temperature modification to the contrastive objective (Eq.~\ref{eqn:info-nce}) to arrive at our proposed pre-training objective:
\begin{equation}\label{eqn:domain-adjust-info-nce}
    \ell_{i} = -log \frac{exp({sim(z_i, z'_i)/\tau_{\alpha}})}{exp({sim(z_i, z'_i)/\tau_{\alpha}}) + \sum_{j=1, j \neq i}^Nexp(sim(z_i, z'_j)/\tau_{ij})}
\end{equation}
The key idea is that for any sample $z_j$ that is likely to be from the same domain as the anchor $z_i$ (i.e., large $w_{ij}$), $\tau_{ij}$ will be small, which acts to increase the relative penalty put on $z_j$ compared to other negative samples. This translates to the network being encouraged to push apart the embeddings of samples that come from the same domain, which helps to avoid capturing the domain structure in the embedding space. As pre-training progresses and domain information is removed, it becomes harder to distinguish between domains up to the limiting case: $\forall i,j: w_{ij} = \frac{1}{N_D} \rightarrow \tau_{ij} = \tau_\alpha$. In other words, the training approaches the case where temperature is fixed for all negative samples -- i.e., the standard contrastive loss of Eq.~\ref{eqn:info-nce}. A gradient analysis of our approach is presented in Appendix~\ref{sec:app_gradient}, and empirical plots of the temperature distributions by pre-training epoch are in Appendix~\ref{sec:app_eff_temp}.

\section{Experiments}\label{sec:Exp}

\textbf{Dataset.} We create an MNIST dataset variant \cite{deng2012mnist} where the digits are colored. We focus on classifying 3 versus 5, which is considered one of the more difficult pairwise tasks. We simulate domain heterogeneity by varying the distribution over digit colors across the domains. For each image in domain $d$, we sample the digit color from $N(\mu_d, \sigma^2)$, where $\mu_d$ is a domain-specific parameter. We create 2 training domains $d=1,2$ with $\mu_1=[255,0,0]$ (red) and $\mu_2=[0,0,255]$ (blue), 1 validation domain $d=3$ with $\mu_3=[255,0,255]$ (purple), and 1 test domain $d=4$ with $\mu_4=[0,255,0]$ (green). We use 23,100 samples ($60\%$ Train, $10\%$ Validation, $10\%$ Test-ID, $20\%$ Test-OOD). Example images are in Appendix~\ref{sec:app_dataset}. By default, we report results as the mean over 5 seeds with $\sigma=50$ and 1\% of digit labels available. 

\textbf{Pre-training and Downstream Evaluation.} We follow the \textit{SimCLR} framework \cite{chen2020simple} for generating positive and negative pairs. 
We use cropping (\textit{RandomResizedCrop()}) and Gaussian blurring data augmentations. After pre-training an encoder $f(x;\theta_f)$ for 400 epochs, we measure its utility by training a linear classifier $h(z;\theta_h)$ on top of the learned embeddings, $z$ (16-dimensional), produced by $f$. We measure performance as the digit classification accuracy of the model $h$ on the test data, reporting results on both the held-out domain (Test-OOD) and a held-out set from the training domains (Test-ID). We also report the accuracy of a domain classifier trained on the learned embeddings to discriminate between the training domains (D-Test-ID). Details on model architecture and hyperparameters are in Appendix~\ref{sec:app_hyperparams}.

\textbf{Baselines.} We compare to \textit{Standard Contrastive Learning} (i.e., Eq.~\ref{eqn:info-nce}) and several domain generalization methods: (i) \textit{Same Domain Negatives} uses only negative samples from the same domain as the anchor, as in \cite{cheng2020subject}; (ii) \textit{Contrastive Learning with MMD Penalty} uses a maximum mean discrepancy (MMD) regularization term to match embedding distributions across domains, as in \cite{li2018domain}; and, (iii) \textit{Contrastive Learning with DANN} uses an adversarial network to match embedding distributions across domains, as in \cite{ganin2016domain}. We also examine standard contrastive learning with \textit{Color Jitter} as a data augmentation; we consider this to be an \textit{oracle} method, since it uses information about domain differences that we often do not have in practice. It gives us an upper bound on expected performance.

\begin{table}  
  \small
  \caption{Test-OOD, Test-ID and D-Test-ID accuracy of methods when evaluated using a downstream linear classifier trained on 1\% of digit labels (i.e., 69 examples). All domains have a color variance of $\sigma=50$. Best digit classification accuracies are \textbf{bolded} and second best accuracies are \underline{underlined}.}
  \vspace{.1cm}
  \label{table:main_results}
  \centering
  \begin{tabular}{lrrr}
    \toprule
    & Test-OOD & Test-ID & D-Test-ID \\
    \midrule
    Standard CL & 0.784 & 0.880 & 0.688 \\
    Same Domain Negatives & 0.728 & 0.896 & 0.496 \\
    CL with MMD Penalty & 0.688 & 0.865 & 0.577 \\
    CL with DANN & 0.700 & 0.846 & 0.543 \\
    Domain-Weighted Negatives (ours) & \underline{0.807} & \underline{0.929} & 0.530 \\
    Domain-Weighted Pairs (ours) & \textbf{0.819} & \textbf{0.945} & 0.520 \\ \midrule
    Standard CL w/Color Jitter (oracle) & 0.978 & 0.976 & 0.490 \\
    \bottomrule
  \end{tabular}
\end{table}

\textbf{Experiment Results.} Table~\ref{table:main_results} shows the digit classification accuracy of each method on Test-ID and Test-OOD data. Our two methods -- \textit{Domain-Weighted Negatives} and \textit{Domain-Weighted Pairs} -- perform best on both sets. To understand the differences in performance, we examine the performance of a linear model trained to classify domains from the embeddings learned by each method. We see that our methods produce among the most domain-invariant embeddings (i.e., lower domain classification accuracy). The only method with lower domain accuracy is \textit{Same Domain Negatives}. Compared to this baseline, our methods have the advantage of being able to learn from negative samples from different domains. We hypothesize this helps them to strike a better balance between learning relevant task information and enforcing domain-invariance. We see that those baselines that are designed to enforce domain invariance (e.g., \textit{CL with MMD penalty}) perform worse than \textit{Standard CL}. A similar finding has been observed in prior work; \cite{gulrajani2020search} found that supervised learning using empirical risk minimization outperforms supervised learning using objectives with a domain invariance penalty. 

\begin{figure}[t]
  \centering
  \includegraphics[width=\linewidth]{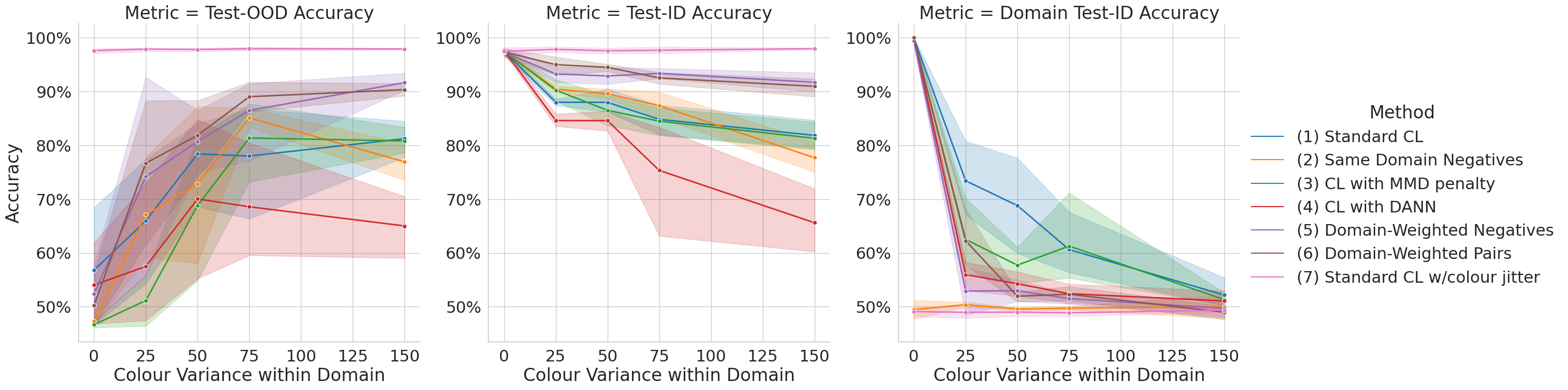}
  \caption{Accuracy of contrastive learning methods as color variance ($\sigma$) within each domain is increased. Downstream linear classifiers are trained using 1\% of digit labels. We see that the proposed domain-weighted methods are frequently among the best performing for Test-OOD Accuracy at $\sigma > 25$. Furthermore, they display higher robustness on Test-ID Accuracy as $\sigma$ is increased.}
  \label{fig:sigma_ablation_plot}
\end{figure}

We examine the robustness of each method to the variance of color ($\sigma$) within domains in Figure~\ref{fig:sigma_ablation_plot}. The proposed domain-weighted methods achieve greater OOD robustness while maintaining ID performance over a broad range of variances; they are frequently among the best performing for Test-OOD Accuracy, and consistently perform best on Test-ID Accuracy. The \textit{Domain Test-ID Accuracy} subplot shows that our proposed methods consistently result in embeddings where it is more difficult to discriminate domains compared to baselines. We conjecture that the greater performance of our methods \textit{in-distribution} as $\sigma$ increases may be explained by their superior ability to ignore domain information -- more variance in color results in more domain information to encode in the finite embedding space. Therefore, methods that ignore this domain information more effectively allow for more capacity to encode digit information.  

\textbf{Label Sparsity, Temperature Hyperparameters \& Model Selection.} Appendix~\ref{app:label-ablation} reports accuracies across downstream label fractions. The benefits of our methods are most pronounced at low label fractions, though they also perform well at higher label fractions (including a fully supervised setting of 100\%). Appendix~\ref{sec:app_temp_ablation} reports the performance of the domain-weighted methods for combinations of $\tau_\alpha$ and $\tau_\beta$ hyperparameters. Appendix~\ref{sec:app_dist_over_hyperparams} shows the distribution in accuracies over all combinations of hyperparamaters. For all methods, there are many combinations with higher accuracy than those selected by our current model selection criterion. 
Model selection is a known challenge in domain generalization \cite{gulrajani2020search}, and alternative strategies should be considered as future work.

\textbf{Conclusion.} We introduce a novel domain-aware loss function to improve the robustness of unsupervised contrastive learning to covariate shift. Preliminary results demonstrate that our method excels at maintaining strong in-distribution performance while also improving out-of-distribution robustness.



\bibliographystyle{ieeetr}
\bibliography{bibliography}


\clearpage
\section*{Supplementary Material}
\renewcommand{\thesubsection}{\Alph{subsection}}

\subsection{Related Work}
\label{sec:app_related_work}

Out of the vast literature on domain generalization (see \cite{gulrajani2020search, zhou2022domain} for a survey), relatively few studies have examined contrastive pre-training strategies. Recent work notes that pre-training with data augmentations more generally is a promising approach to improve distribution shift robustness \cite{wiles2022} over conventional domain generalisation baselines. Shen et al. \cite{shen2022connect} demonstrates the utility of contrastive pre-training for unsupervised domain adaptation, a related, but different problem, in which unlabeled data from the test domain are available while pre-training. Closest to our work are \cite{harary2022unsupervised} and \cite{zhang2022towards}, which both study the problem of domain generalization from sparsely labelled data and propose strategies based on contrastive pre-training. However, our work resolves some key limitations with each of these. In particular, \cite{harary2022unsupervised} is only relevant to image data and relies on the existence of a known mapping function that can be used to reduce differences between domains (e.g., they use an edge detection model). Similar to our approach, \cite{zhang2022towards} proposes to use domain dependent weights on the negative samples in a batch, but unlike our method, their weights are fixed and do not adapt with the changing state of the learned embedding space. Furthermore, they do not adjust temperature as a means of controlling the weighting; instead, they weight the negative samples outside of the exponent, giving their method different mathematical properties. Finally, they use domain-specific queues of negative samples, which may be challenging to maintain with increasing numbers of domains.

Separately, recent works have considered varying the temperature during contrastive pre-training: \cite{kukleva2023temperature} modify temperature on a cosine schedule, while \cite{pmlr-v202-qiu23a,pmlr-v202-huang23c} adaptively vary temperature on a pairwise basis based on properties of the learned embedding space. However, all of these works are focused on modifying temperature to address label imbalance; by contrast, our work focuses on modifying temperature to improve covariate shift robustness.

\subsection{Gradient Analysis of Proposed Approach}
\label{sec:app_gradient}

Inspired by \cite{wang2021understanding}, we analyze the gradients with respect to different negative samples. This allows us to show how our method influences the distribution of gradients, and thereby influences what attributes of the data are learned versus not learned. For ease of notation, let $s_{ij} = sim(z_i, z_j)$ represent the cosine similarity between embeddings $z_i$ and $z_j$. The contribution of negative pair similarity $s_{ij}$ to the update of encoder parameters $\theta_f$ is weighted by the partial derivative of loss $\ell_i$ with respect to $s_{ij}$, which can be computed as:
\begin{align*}
    \frac{\partial \ell_i}{\partial s_{ij}} &= \frac{exp(s_{ij}/\tau_{ij})(1/\tau_{ij})}{exp(s_{ii}/\tau_{\alpha}) + \sum_{k\neq i} exp(s_{ik}/\tau_{ik})}
\end{align*}
Since the denominator is constant for all $s_{ij}$, we get that the relative weight for $s_{ij}$ is:
\begin{align*}
    \frac{\partial \ell_i}{\partial s_{ij}} \propto exp(s_{ij}/\tau_{ij})(1/\tau_{ij})
\end{align*}
As in standard contrastive learning (i.e., fixed temperature), the more similar a negative sample is to the anchor (i.e., larger $s_{ij}$), the larger its relative weight; this means that the loss is \textit{hardness-aware} \cite{wang2021understanding}. However, in our approach, since the temperature is now sample-dependent, the relative penalty is also influenced by the temperature. As a result, a higher penalty is given to negative samples that are more likely to come from the same domain as the anchor, since they have a small temperature $\tau_{ij}$. This analysis also reveals why it is useful to modify the temperature parameter as a means of controlling the relative weighting of negatives; if we instead included a weighting term outside of the exponent, its effect would be negligible compared to the scaling of the pairwise cosine similarities.

The gradient our proposed loss function \ref{eqn:domain-adjust-info-nce} with respect to encoder parameters $\theta_f$ can be written as:
\begin{align*}
    \nabla_{\theta_f} \ell_i = \sum_{j=1}^N \frac{\partial \ell_i}{\partial s_{ij}} \nabla_{\theta_f} s_{ij}
\end{align*}
using the chain rule. Therefore, the contribution of negative pair similarity $s_{ij}$ to the update of $\theta_f$ is weighted by $\frac{\partial \ell_i}{\partial s_{ij}}$.

\subsection{Specifying and Training the Domain Discriminator}
\label{sec:app_train_domain_discriminator}

As outlined in Section~\ref{sec:method}, at each pre-training epoch we train a domain discriminator network $g: \mathbb{R}^k \rightarrow \mathbb{R}^{|D|}$ on the output embeddings $z_i$ of the encoder network and then use the probabilistic output of this discriminator ($P(D = d | z_i) = softmax(g(z_i);\theta_g)$) to weight negative pairs in Eq.~\ref{eqn:domain-adjust-info-nce}. The domain discriminator is trained on the pre-training data, and then used to apply probabilistic weights onto the same pre-training data. 

There are several design decisions to consider when specifying the domain discriminator. First, either a linear or a non-linear classifier can be used. In these experiments we opt for a linear classifier though future work will consider increasing the representational capacity of the domain discriminator. Second, the domain discriminator can be fit at either the \textit{global-level} -- i.e., a single domain discriminator is fit on all instances at the beginning of each epoch -- or the \textit{batch-level} -- i.e., a new domain discriminator is fit on each batch of pre-training data. The \textit{batch-level} approach is designed to break up more local domain heterogeneity in the data (see \cite{Zhu_2022_CVPR} for a related approach in a supervised DANN architecture). Rather than fitting a single decision boundary through all pre-training instances, it has the flexibility to fit a decision-boundary only for the K instances in each batch. Over the course of training, membership of instances in each batch changes at random, and thus the encoder is consistently pushed to break up any local domain heterogeneity in the embedding space. We consider a \textit{global-level} domain discriminator in our main experiments but we ablate this with \textit{batch-level} discriminators in Appendix~\ref{sec:app_global_vs_batch}. Finally, one could also further refine the ability of the domain discriminator to break up domain heterogeneity by investigating how well-calibrated its output distribution is $P(D = d | z_i)$. This investigation is left as important future work.

\subsection{Additional Experiment Settings, Model Architecture \& Hyperparameters}
\label{sec:app_hyperparams}

We implement the models in PyTorch. All methods are pre-trained for 400 epochs. We use the Adam optimizer with an initial learning rate of 0.001 and a learning rate scheduler with step size of 20 and $\gamma = 0.9$.

We use the same model architecture for all methods. The encoder network is a convolutional neural network (CNN), consisting of three convolutional blocks each with batch normalization, dropout and ReLU activation, followed by a single linear layer. In all cases where a domain discriminator network is used, we implement it as a linear classifier. We work with 16-dimensional embeddings, $z$, extracted directly from the encoder and do \textbf{not} use a projector head in our experiments. The decision to not use a projector head was made following preliminary experiments with the \textit{Standard Contrastive Learning} model on the dataset, in which the use of a linear or 2-layer MLP projector head made both Test-ID and Test-OOD results substatnially worse. More details for this are provided in Appendix~\ref{sec:app_projector_head}.

Regarding data augmentation, we use \textit{Resized Random Crop} and \textit{Gaussian Blur} as defaults in all models. While SimCLR \cite{chen2020simple} also uses \textit{Random Horizontal Flip} and \textit{Color Jittering} in their default set, we do not. First, as we are working with digits 3 and 5, we do not use \textit{Random Horizontal Flip} as it would not be label-preserving, and thus would hurt performance. Second, as digit color is mediated by the domain, color jittering trivially removes domain-variant information in this toy dataset. Therefore, we include a contrastive learning model with color jittering as an \textit{oracle}.

In our main experiments, we select hyperparameters based on classification accuracy on data from the validation domain at a 1\% label fraction. We sweep over the following hyperparameter ranges and select the hyperaparameters for each method with the best average validation set accuracy over 5 random seeds of the data
\begin{itemize}
    \item \textit{Standard Contrastive Learning}, \textit{Same Domain Negatives} \& \textit{Standard CL w/Color Jitter}: we sweep over temperatures in the range $\tau \in \{0.05, 0.075, 0.1, 0.125, 0.15, 0.175, 0.2, 0.25, 0.5, 1.0\}$ 
    \item \textit{Domain-Weighted Negatives} and \textit{Domain-Weighted Pairs}: we sweep over temperatures in the range $\tau_\alpha \in \{0.05, 0.075, 0.1, 0.125, 0.15, 0.175, 0.2, 0.25, 0.5, 1.0\}$ and $\tau_\beta \in \{0.1, 0.25, 0.5, 1.0\}$ 
    \item \textit{CL with MMD penalty} and \textit{CL with DANN}: we sweep over temperatures in the range $\tau \in \{0.05, 0.075, 0.1, 0.125, 0.15, 0.175, 0.2, 0.25, 0.5, 1.0\}$ and penalty weights in the range $\lambda \in \{0.01, 0.1, 1.0, 5.0\}$
\end{itemize}

\textbf{Note:} we select hyperparameters separately at each value of $\sigma$ in Figure~\ref{fig:sigma_ablation_plot}.

\subsection{Dataset Details}
\label{sec:app_dataset}

\begin{figure}
     \centering
     \begin{subfigure}[b]{0.22\textwidth}
         \centering
         \includegraphics[width=\textwidth]{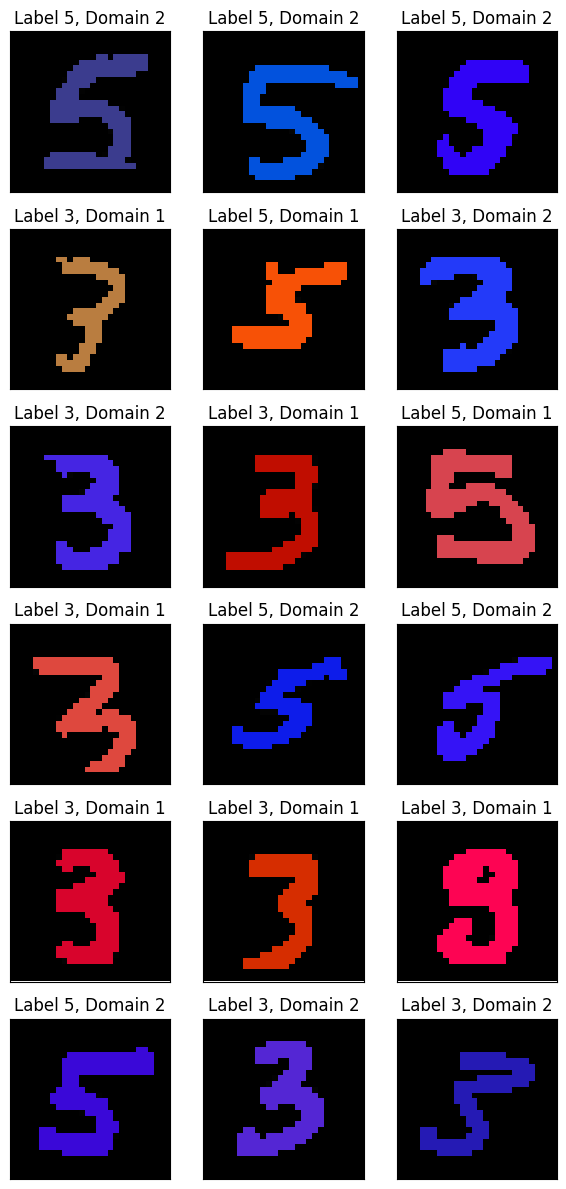}
         \caption{Train}
     \end{subfigure}
     \hfill
     \begin{subfigure}[b]{0.22\textwidth}
         \centering
         \includegraphics[width=\textwidth]{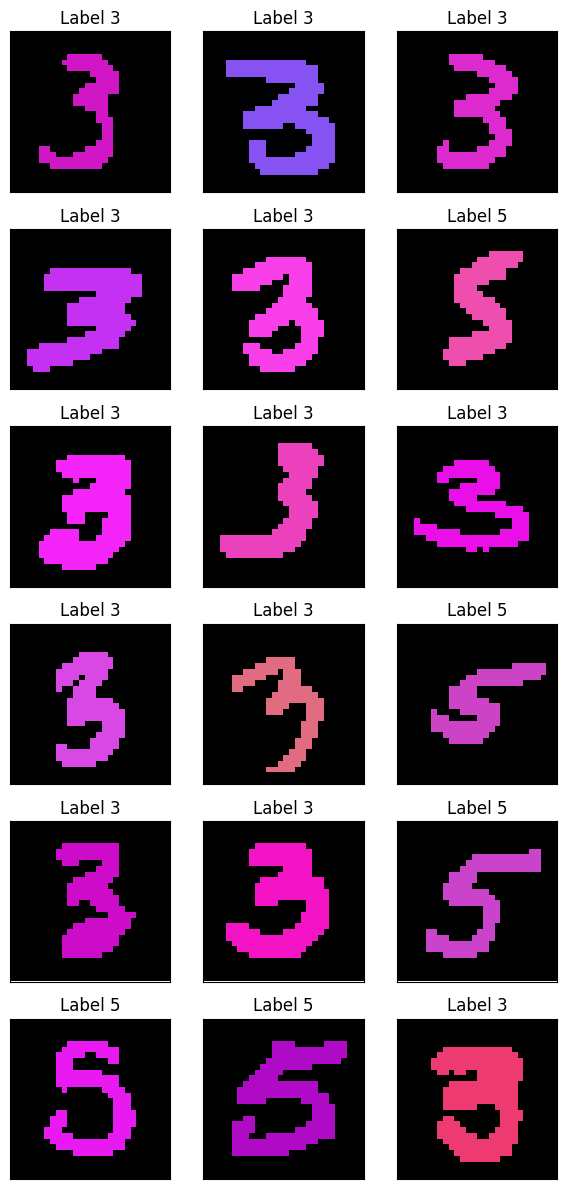}
         \caption{Validation}
     \end{subfigure}
     \hfill
     \begin{subfigure}[b]{0.22\textwidth}
         \centering
         \includegraphics[width=\textwidth]{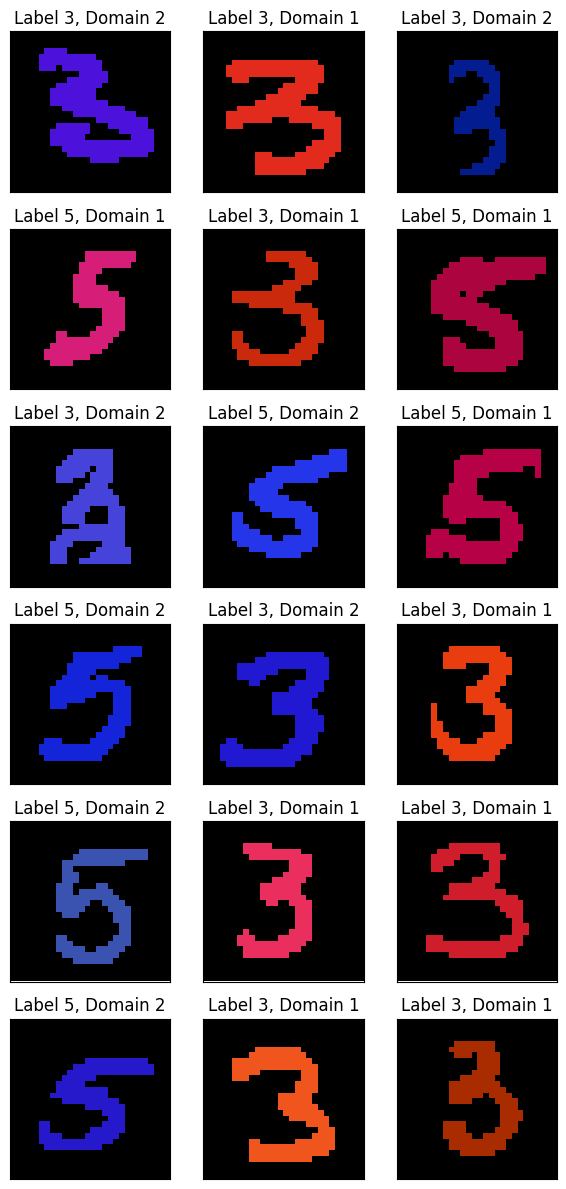}
         \caption{Test-ID}
     \end{subfigure}
     \hfill
     \begin{subfigure}[b]{0.22\textwidth}
         \centering
         \includegraphics[width=\textwidth]{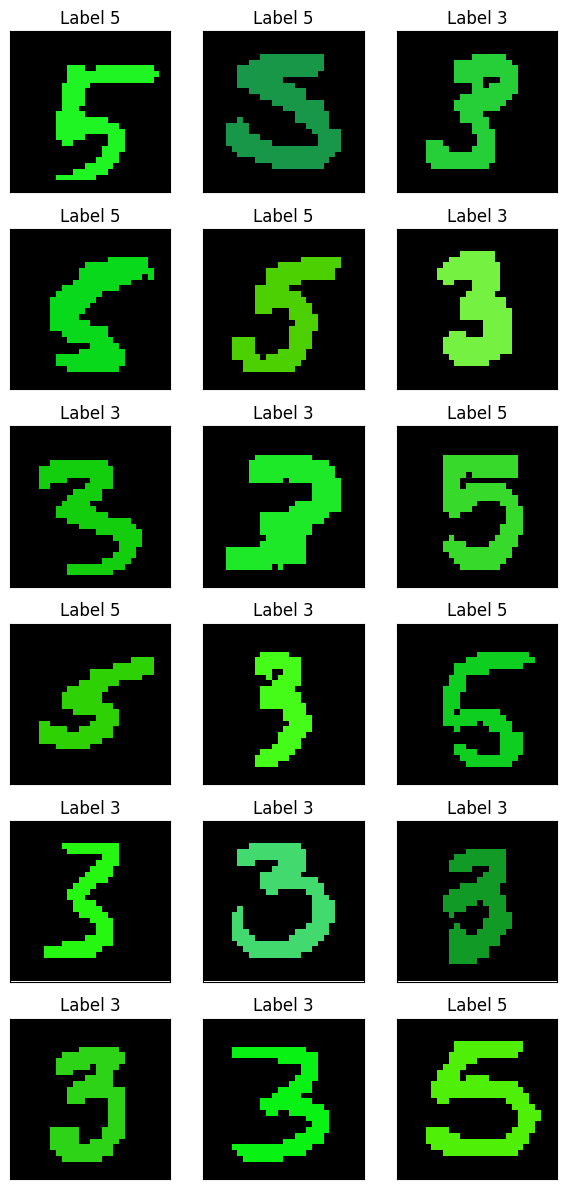}
         \caption{Test-OOD}
     \end{subfigure}
        \caption{Randomly sampled example digits from the Colored-MNIST dataset variant that we use for the experiments in this paper. Here the colors of the digits are sampled around the mean pixel value for each domain with a variance of $\sigma = 50$. We show examples from each of the dataset splits.}
        \label{fig:example_digits}
\end{figure}

Figure~\ref{fig:example_digits} shows example digits from each of the different dataset splits. The digits are colored but the background is left black in all images. We create 2 training domains $d=1,2$ with $\mu_1=[255,0,0]$ (red) and $\mu_2=[0,0,255]$ (blue), 1 validation domain $d=3$ with $\mu_3=[255,0,255]$ (purple), and 1 test domain $d=4$ with $\mu_4=[0,255,0]$ (green). There is no variation in color \textit{within} each image, but there is variance in color \textit{between} images from the same domain (controlled by the parameter $\sigma$). There are 11,548 digits total in the dataset, which we split into disjoint sets: training has 6,930 images, validation has 1,154, Test-ID has 1,154, and Test-OOD has 2,310. For the label sparsity ablations, 1\% of digit labels corresponds to 69 instances, 10\% to 693, and 100\% to 6,930.  

\begin{figure}
     \centering
     \begin{subfigure}[b]{0.45\textwidth}
         \centering
         \includegraphics[width=\textwidth]{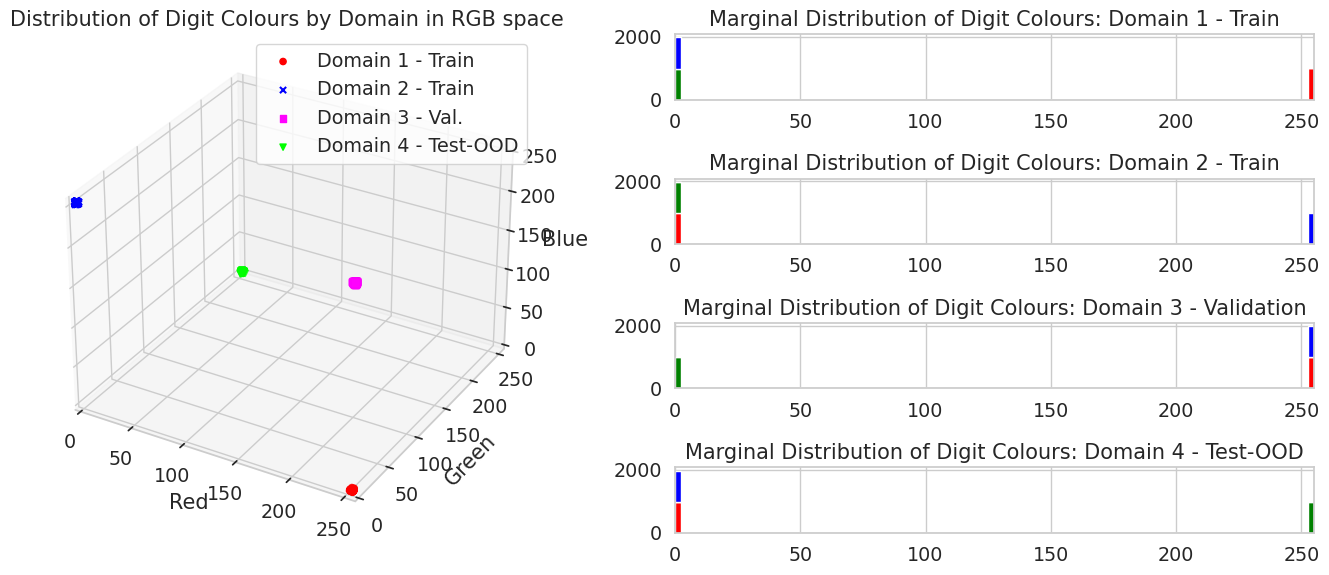}
         \caption{$\sigma=0$}
     \end{subfigure}
     \hfill
     \begin{subfigure}[b]{0.45\textwidth}
         \centering
         \includegraphics[width=\textwidth]{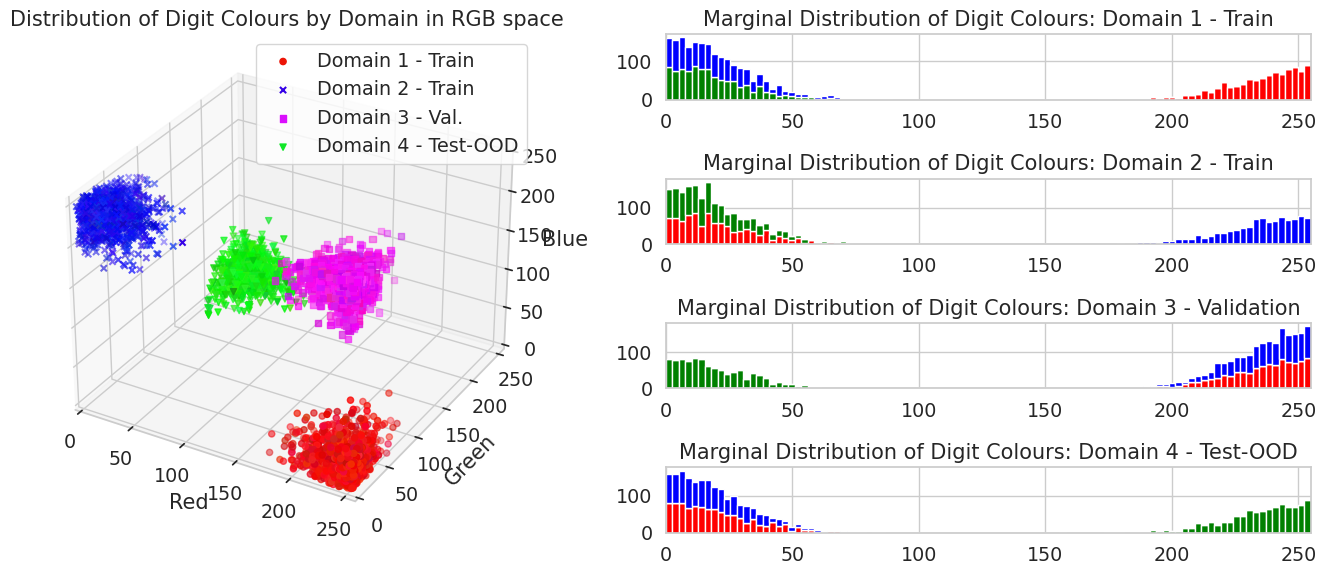}
         \caption{$\sigma=25$}
     \end{subfigure}
     \par\bigskip
     \begin{subfigure}[b]{0.45\textwidth}
         \centering
         \includegraphics[width=\textwidth]{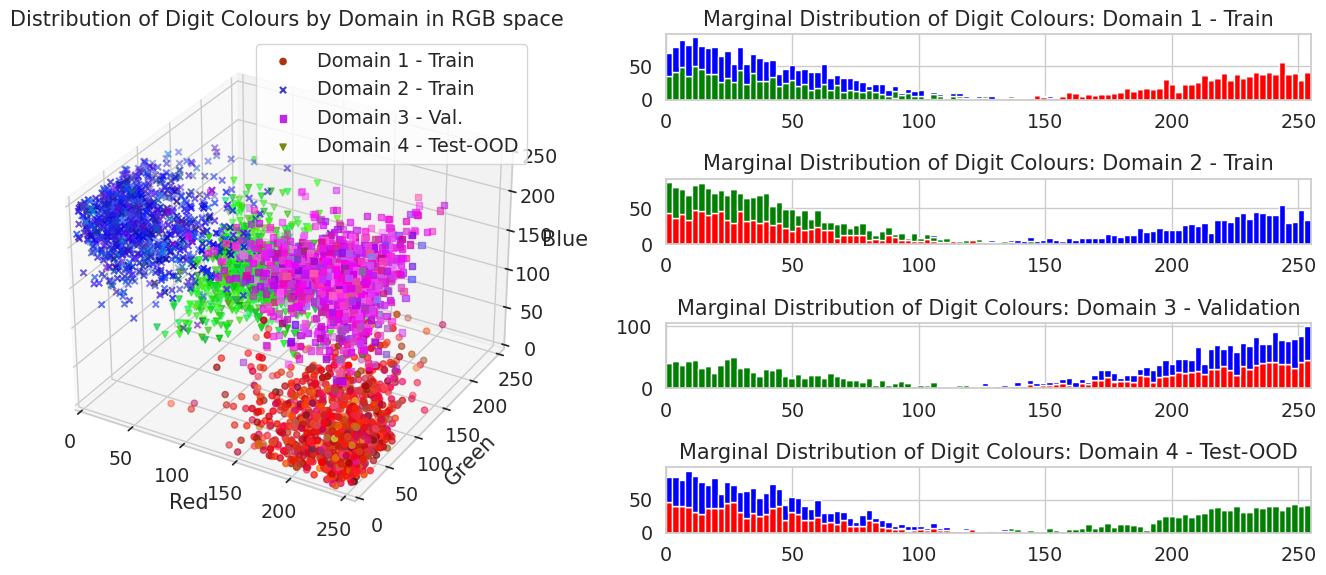}
         \caption{$\sigma=50$}
     \end{subfigure}
     \hfill
     \begin{subfigure}[b]{0.45\textwidth}
         \centering
         \includegraphics[width=\textwidth]{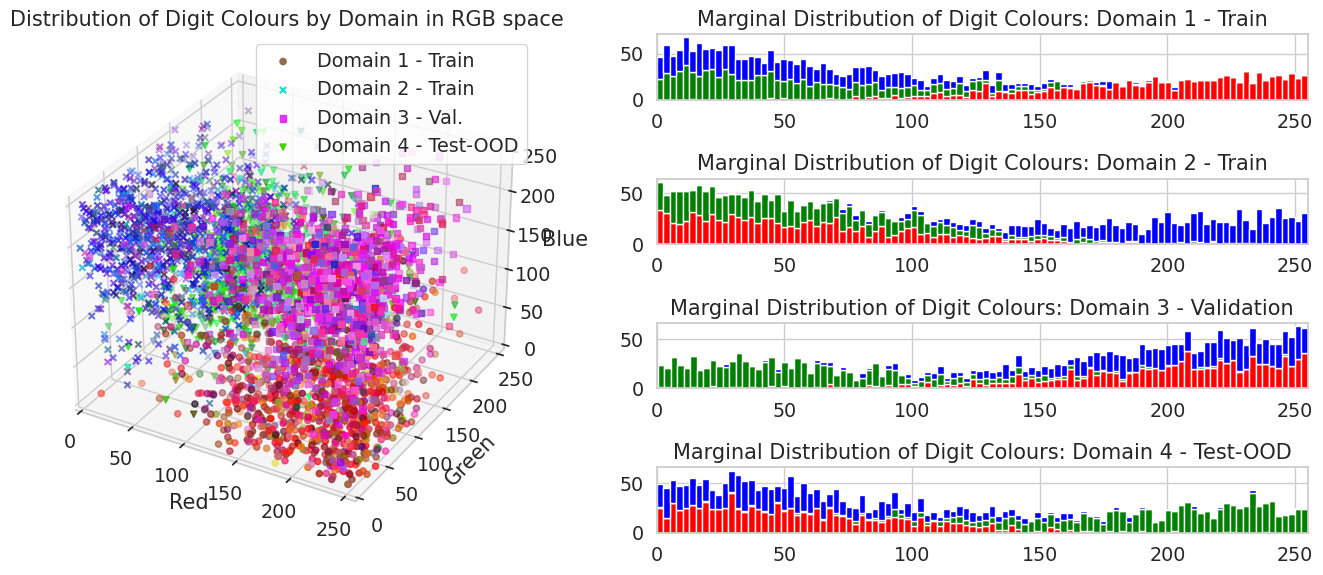}
         \caption{$\sigma=75$}
    \end{subfigure}
    \par\bigskip
    \begin{subfigure}[b]{0.45\textwidth}
         \centering
         \includegraphics[width=\textwidth]{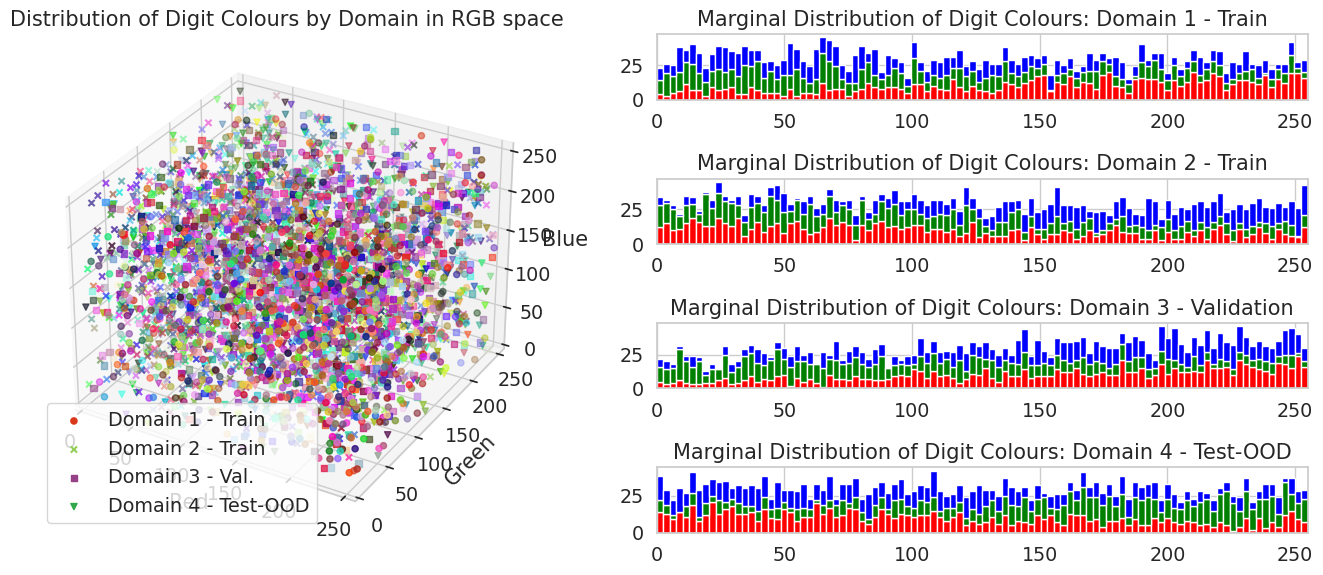}
         \caption{$\sigma=150$}
    \end{subfigure}
    \hfill
    \caption{The left-hand side of each subplot shows the distribution of digit colors in RGB-space; the right hand side shows the marginal distribution of their intensities by each color dimension. As $\sigma$ increases there is more color variance within each of the domains, to the extent that domains start overlapping at high $\sigma$.}
    \label{fig:color_distributions}
\end{figure}

Figure~\ref{fig:color_distributions} shows how the colors of the digits within each domain vary as $\sigma$ is increased. As $\sigma$ increases there is more color variance within each of the domains, to the extent that domains start overlapping at high $\sigma$. 

\subsection{Additional Experimental Results}
\label{sec:app_add_results}

\subsubsection{Label Ablation Table}\label{app:label-ablation}

Table~\ref{table:label_ablation} shows performance of the proposed domain-weighted methods relative to baselines at different downstream digit label fractions. We see that advantage of our proposed methods are most pronounced at low label fractions (e.g., 1\%) but that they remain competitive with baselines at higher label fractions. Notably, they always achieve the highest Test-ID accuracy regardless of label fraction.

\begin{table}[h]
    \caption{Test-OOD and Test-ID accuracy of contrastive learning methods when evaluated using a downstream logistic regression classifier trained on different fractions of digit labels. All domains have a color variance of $\sigma=50$. Best accuracies are \textbf{bolded} and second best are \underline{underlined}.}
    \vspace{2mm}
    \label{table:label_ablation}
    \centering
    \begin{tabular}{lrrrrrr}
        \toprule
        & \multicolumn{3}{r}{Test-OOD Accuracy} & \multicolumn{3}{r}{Test-ID Accuracy} \\
        \cmidrule(r){2-4}\cmidrule(r){5-7}
        Label Fraction & 1\% & 10\% & 100\% & 1\% & 10\% & 100\% \\
        \midrule
        Standard CL & 0.784 & 0.774 & 0.716 & 0.880 & 0.933 & 0.937 \\
        Same Domain Negatives & 0.728 & 0.856 & \underline{0.886} & 0.896 & 0.956 & 0.960 \\
        CL with MMD penalty & 0.688 & \textbf{0.908} & 0.826 & 0.865 & 0.941 & 0.946 \\
        CL with DANN & 0.700 & 0.752 & 0.764 & 0.846 & 0.917 & 0.922 \\
        Domain-Weighted Negatives (ours) & \underline{0.807} & 0.899 & \textbf{0.908} & \underline{0.929} & \underline{0.963} & \textbf{0.967} \\
        Domain-Weighted Pairs (ours) & \textbf{0.819} & \underline{0.903} & 0.871 & \textbf{0.945} & \textbf{0.963} & \underline{0.964} \\
        \midrule
        Standard CL w/Color Jitter (oracle) & 0.978 & 0.985 & 0.985 & 0.976 & 0.983 & 0.986 \\
        \bottomrule
    \end{tabular}
\end{table}

\subsubsection{Domain-Weighted Methods Performance by Hyperparameters}
\label{sec:app_temp_ablation}

Tables~\ref{table:dw_negatives_temp_val}-\ref{table:dw_pairs_temp_test_id} show the performance of the proposed methods by temperature hyperparameter values. The distribution in results is intuitive, with the models with a lower $\tau_\alpha$ typically performing best (higher weight applied to hard negatives). However, we also see that models selected per validation set accuracy do not necessarily correspond to the highest average test set accuracies, suggesting that alternative model selection strategies might yield further improvements in performance.

\subsubsection{Distribution of Accuracies over all Hyperparameter Values}
\label{sec:app_dist_over_hyperparams}

\begin{figure}[t]
  \centering
  \includegraphics[width=0.9\linewidth]{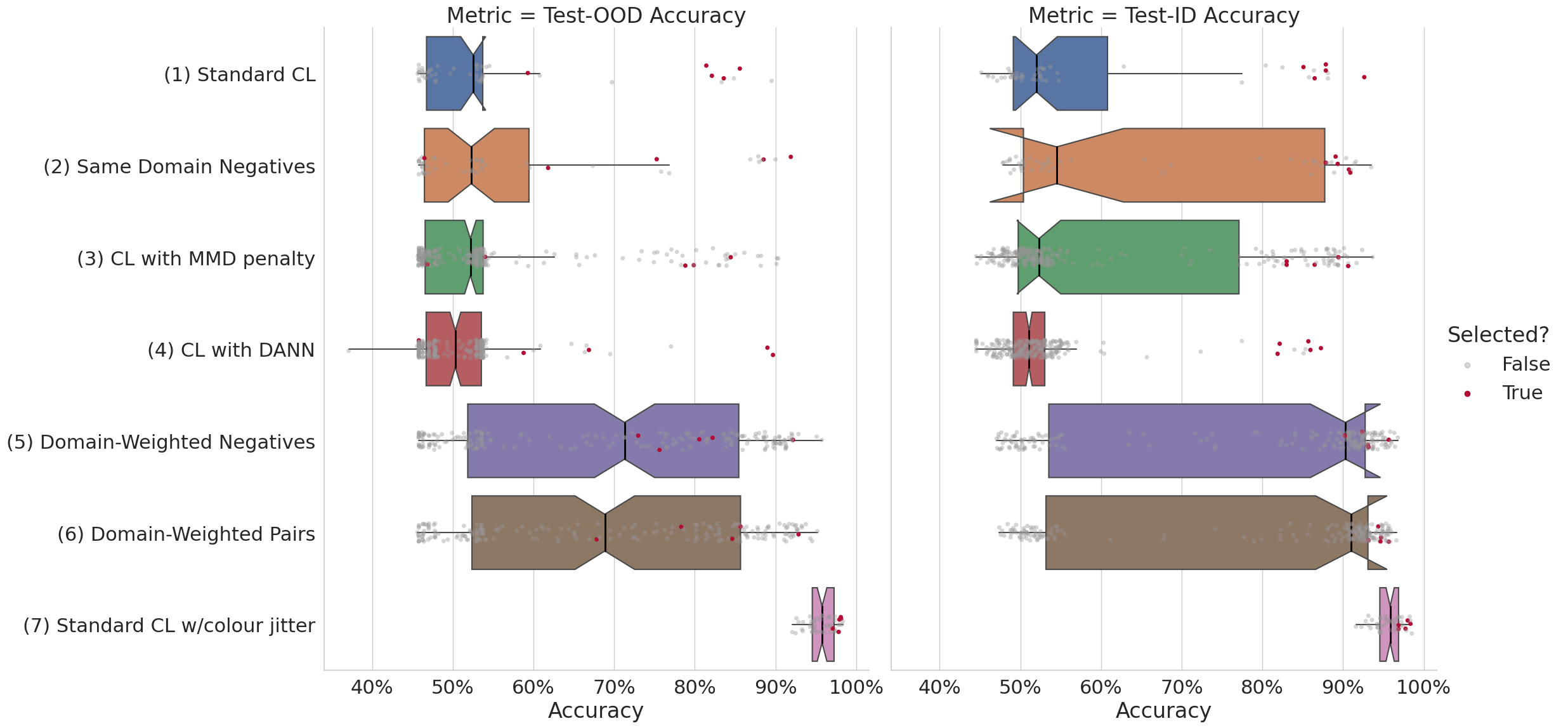}
  \caption{Accuracy of all trained models at color variance $\sigma = 50$ and with downstream linear classifiers trained using 1\% of digit labels. Red circles represent the models selected using the strategy of highest validation accuracy on the held-out training domain (where accuracy is averaged over 5 random seeds); grey circles are the other models trained but not selected.}
  \label{fig:model_selection_plot}
\end{figure}

Figure~\ref{fig:model_selection_plot} shows the distribution of model performance over all hyperparameter combinations (the models selected by validation accuracy on the held-out training domain are highlighted in red; hyperparameters considered are documented in Appendix~\ref{sec:app_hyperparams}). First, we see that our proposed methods have higher performance on average relative to baselines on both the Test-OOD and Test-ID accuracy metrics, where the former is particularly pronounced. This may suggest our proposed methods are more robust to the choice of hyperparameters, which may be a useful property when scaling to more complex datasets. However, we also see that there are many high performing models on Test-OOD accuracy (and to a lesser degree on Test-ID accuracy) that are not selected by the model selection strategy. This appears to affect all methods that attempt to correct for domain heterogeneity, yet \textit{Standard Contrastive Learning} appears relatively unencumbered by it. Understanding the reason for this incongruence between validation and test accuracy is important future work and -- given there is a high density of trained models from our proposed method that achieve accuracies $>90\%$ -- it is plausible that alternative model selection strategies (e.g., selecting hyperparameters on a different metric or on different folds of validation data) could result in an even better performing set of hyperparameters being selected.    

\begin{table}[h]
    \caption{Temperature ablation for \textbf{domain-weighted negatives} method at 1\% digit label fraction and domain color variance $\sigma = 50$ -- Validation Accuracy.}
    \vspace{2mm}
    \label{table:dw_negatives_temp_val}
    \centering
    \begin{tabular}{lrrrr}
        \toprule
        $\tau_\alpha \downarrow$, $\tau_\beta \rightarrow$ & 0.1 & 0.25 & 0.5 & 1.0 \\
        \midrule
        0.05 & 0.909 & 0.918 & 0.894 & 0.884 \\
        0.075 & 0.889 & 0.887 & \textbf{0.923} & 0.906 \\
        0.1 & 0.875 & 0.902 & 0.885 & 0.919 \\
        0.125 & 0.845 & 0.908 & 0.895 & 0.910 \\
        0.15 & 0.741 & 0.856 & 0.891 & 0.907 \\
        0.175 & 0.484 & 0.830 & 0.847 & 0.901 \\
        0.2 & 0.499 & 0.594 & 0.851 & 0.903 \\
        0.25 & 0.498 & 0.492 & 0.736 & 0.830 \\
        0.5 & 0.503 & 0.500 & 0.506 & 0.588 \\
        1.0 & 0.510 & 0.514 & 0.517 & 0.511 \\
        \bottomrule
    \end{tabular}
\end{table}

\begin{table}[h]
    \caption{Temperature ablation for \textbf{domain-weighted negatives} method at 1\% digit label fraction and domain color variance $\sigma = 50$ -- Test-OOD Accuracy. Models with best accuracies are \textbf{bolded} though note these do not necessarily correspond to the model selected on validation data (* prefix).}
    \vspace{2mm}
    \label{table:dw_negatives_temp_test_ood}
    \centering
    \begin{tabular}{lrrrr}
        \toprule
        $\tau_\alpha \downarrow$, $\tau_\beta \rightarrow$ & 0.1 & 0.25 & 0.5 & 1.0 \\
        \midrule
        0.05 & 0.738 & 0.753 & 0.841 & 0.841 \\
        0.075 & 0.724 & 0.737 & *0.807 & 0.855 \\
        0.1 & 0.692 & 0.859 & 0.870 & \textbf{0.883} \\
        0.125 & 0.663 & 0.862 & 0.805 & 0.837 \\
        0.15 & 0.573 & 0.750 & 0.789 & 0.866 \\
        0.175 & 0.477 & 0.710 & 0.793 & 0.794 \\
        0.2 & 0.511 & 0.494 & 0.746 & 0.790 \\
        0.25 & 0.505 & 0.497 & 0.660 & 0.731 \\
        0.5 & 0.515 & 0.490 & 0.513 & 0.501 \\
        1.0 & 0.510 & 0.507 & 0.499 & 0.481 \\
        \bottomrule
    \end{tabular}
\end{table}

\begin{table}[h]
    \caption{Temperature ablation for \textbf{domain-weighted negatives} method at 1\% digit label fraction and domain color variance $\sigma = 50$ -- Test-ID Accuracy. Models with best accuracies are \textbf{bolded} though note these do not necessarily correspond to the model selected on validation data (* prefix).}
    \vspace{2mm}
    \label{table:dw_negatives_temp_test_id}
    \centering
    \begin{tabular}{lrrrr}
        \toprule
        $\tau_\alpha \downarrow$, $\tau_\beta \rightarrow$ & 0.1 & 0.25 & 0.5 & 1.0 \\
        \midrule
        0.05 & 0.910 & 0.931 & 0.924 & 0.913 \\
        0.075 & 0.921 & 0.936 & *0.929 & 0.930 \\
        0.1 & 0.923 & 0.921 & 0.941 & 0.940 \\
        0.125 & 0.877 & 0.932 & 0.928 & \textbf{0.949} \\
        0.15 & 0.759 & 0.907 & 0.913 & 0.945 \\
        0.175 & 0.508 & 0.885 & 0.892 & 0.936 \\
        0.2 & 0.499 & 0.662 & 0.875 & 0.928 \\
        0.25 & 0.495 & 0.503 & 0.753 & 0.886 \\
        0.5 & 0.499 & 0.498 & 0.500 & 0.575 \\
        1.0 & 0.508 & 0.511 & 0.511 & 0.506 \\
        \bottomrule
    \end{tabular}
\end{table}

\begin{table}[h]
    \caption{Temperature ablation for \textbf{domain-weighted pairs} method at 1\% digit label fraction and domain color variance $\sigma = 50$ -- Validation Accuracy.}
    \vspace{2mm}
    \label{table:dw_pairs_temp_val}
    \centering
    \begin{tabular}{lrrrr}
        \toprule
        $\tau_\alpha \downarrow$, $\tau_\beta \rightarrow$ & 0.1 & 0.25 & 0.5 & 1.0 \\
        \midrule
        0.05 & 0.917 & 0.885 & 0.923 & 0.919 \\
        0.075 & 0.896 & 0.916 & 0.930 & 0.905 \\
        0.1 & 0.888 & 0.926 & 0.920 & 0.924 \\
        0.125 & 0.817 & 0.928 & 0.935 & 0.903 \\
        0.15 & 0.688 & 0.912 & 0.928 & 0.935 \\
        0.175 & 0.501 & 0.887 & 0.936 & \textbf{0.948} \\
        0.2 & 0.497 & 0.544 & 0.914 & 0.924 \\
        0.25 & 0.494 & 0.496 & 0.892 & 0.889 \\
        0.5 & 0.508 & 0.494 & 0.504 & 0.587 \\
        1.0 & 0.528 & 0.519 & 0.514 & 0.513 \\
        \bottomrule
    \end{tabular}
\end{table}

\begin{table}[h]
    \caption{Temperature ablation for \textbf{domain-weighted pairs} method at 1\% digit label fraction and domain color variance $\sigma = 50$ -- Test-OOD Accuracy. Models with best accuracies are \textbf{bolded} though note these do not necessarily correspond to the model selected on validation data (* prefix).}
    \vspace{2mm}
    \label{table:dw_pairs_temp_test_ood}
    \centering
    \begin{tabular}{lrrrr}
        \toprule
        $\tau_\alpha \downarrow$, $\tau_\beta \rightarrow$ & 0.1 & 0.25 & 0.5 & 1.0 \\
        \midrule
        0.05 & 0.638 & 0.726 & 0.788 & 0.730 \\
        0.075 & 0.780 & 0.776 & 0.712 & 0.822 \\
        0.1 & 0.703 & 0.717 & 0.789 & 0.849 \\
        0.125 & 0.654 & 0.818 & 0.794 & 0.795 \\
        0.15 & 0.501 & 0.847 & 0.797 & 0.855 \\
        0.175 & 0.482 & 0.708 & 0.861 & *0.819 \\
        0.2 & 0.519 & 0.478 & \textbf{0.881} & 0.755 \\
        0.25 & 0.486 & 0.499 & 0.860 & 0.836 \\
        0.5 & 0.517 & 0.509 & 0.496 & 0.738 \\
        1.0 & 0.504 & 0.482 & 0.496 & 0.494 \\
        \bottomrule
    \end{tabular}
\end{table}

\begin{table}[h]
    \caption{Temperature ablation for \textbf{domain-weighted pairs} method at 1\% digit label fraction and domain color variance $\sigma = 50$ -- Test-ID Accuracy. Models with best accuracies are \textbf{bolded} though note these do not necessarily correspond to the model selected on validation data (* prefix).}
    \vspace{2mm}
    \label{table:dw_pairs_temp_test_id}
    \centering
    \begin{tabular}{lrrrr}
        \toprule
        $\tau_\alpha \downarrow$, $\tau_\beta \rightarrow$ & 0.1 & 0.25 & 0.5 & 1.0 \\
        \midrule
        0.05 & 0.918 & 0.910 & 0.929 & 0.933 \\
        0.075 & 0.931 & 0.927 & 0.925 & 0.923 \\
        0.1 & 0.920 & 0.928 & 0.933 & 0.931 \\
        0.125 & 0.863 & 0.924 & 0.937 & 0.928 \\
        0.15 & 0.739 & 0.924 & 0.931 & 0.940 \\
        0.175 & 0.502 & 0.873 & 0.925 & *\textbf{0.945} \\
        0.2 & 0.511 & 0.532 & 0.922 & 0.940 \\
        0.25 & 0.508 & 0.511 & 0.872 & 0.927 \\
        0.5 & 0.508 & 0.518 & 0.510 & 0.642 \\
        1.0 & 0.518 & 0.520 & 0.516 & 0.516 \\
        \bottomrule
    \end{tabular}
\end{table}

\subsubsection{Global vs. Batch-level Domain Discriminator in Domain-Weighted Methods}
\label{sec:app_global_vs_batch}

As discussed in Appendix~\ref{sec:app_train_domain_discriminator}, our proposed methods allow for the domain discriminator to be fit at either the \textit{global-level} (all pre-training instances) or the \textit{batch-level}. Table~\ref{tab:global_vs_batch} ablates these approaches. While there is some variation in the performance of the different methods, we do not yet see a clear trend the supports the use of one approach over the other. It is plausible that the MNIST variant we are using in these experiments does not exhibit much substructure in the domain heterogeneity it manifests, and thus either a \textit{global-level} or a \textit{batch-level} discriminator is effective at estimating how domains are separated.  

\begin{table}[h] \tiny
    \caption{Comparison of domain-weighted methods where the domain discriminator used to apply the weighting to the negative pairs is trained at either the global (i.e., all instances) or the batch-level (i.e., retrained for each batch of data). All results are at the 1\% label fraction and models were selected using the validation data. Best accuracies are \textbf{bolded} for each variant of the loss function.}
    \vspace*{2mm}
    \label{tab:global_vs_batch}
    
    \begin{subtable}{\linewidth}
        \centering
        \begin{tabular}{lrrr|rrr|rrr}
            \toprule
            & \multicolumn{3}{r|}{$\sigma = 0$} & \multicolumn{3}{r|}{$\sigma = 25$} & \multicolumn{3}{r}{$\sigma = 50$} \\
            & Test-OOD & Test-ID & D-Test-ID & Test-OOD & Test-ID & D-Test-ID & Test-OOD & Test-ID & D-Test-ID \\
            \midrule
            Domain-Weighted Negatives - global & \textbf{0.524} & 0.973 & 0.994 & 0.742 & 0.933 & 0.529 & \textbf{0.807} & 0.929 & 0.530 \\
            Domain-Weighted Negatives - batch & 0.467 & \textbf{0.974 }& 0.502 & \textbf{0.769} & \textbf{0.956} & 0.509 & 0.775 & \textbf{0.931} & 0.500 \\
            \midrule
            Domain-Weighted Pairs - global & 0.503 & 0.974 & 1.000 & 0.767 & \textbf{0.950} & 0.623 & \textbf{0.819} & \textbf{0.945} & 0.520 \\
            Domain-Weighted Pairs - batch & \textbf{0.521} & \textbf{0.976} & 1.000 & \textbf{0.779} & 0.934 & 0.564 & 0.806 & 0.928 & 0.535 \\
            \bottomrule
        \end{tabular}
    \end{subtable}
    \begin{subtable}{\linewidth}
        \centering
        \begin{tabular}{lrrr|rrr}
            \toprule
            & \multicolumn{3}{r|}{$\sigma = 75$} & \multicolumn{3}{r}{$\sigma = 150$} \\
            & Test-OOD & Test-ID & D-Test-ID & Test-OOD & Test-ID & D-Test-ID \\
            \midrule
            Domain-Weighted Negatives - global & 0.865 & \textbf{0.934} & 0.515 & 0.917 & 0.917 & 0.497 \\
            Domain-Weighted Negatives - batch & \textbf{0.913} & 0.925 & 0.503 & \textbf{0.928} & \textbf{0.932} & 0.506 \\
            \midrule
            Domain-Weighted Pairs - global & \textbf{0.890} & 0.925 & 0.523 & 0.904 & 0.910 & 0.490 \\
            Domain-Weighted Pairs - batch & 0.887 & \textbf{0.927} & 0.511 & \textbf{0.920} & \textbf{0.920} & 0.498 \\
            \bottomrule
        \end{tabular}
    \end{subtable}

\end{table}

\clearpage \newpage

\subsection{Projector Head Ablation Results with Standard Contrastive Learning}
\label{sec:app_projector_head}

\begin{figure}
     \centering
     \begin{subfigure}[b]{0.225\textwidth}
         \centering
         \includegraphics[width=\textwidth]{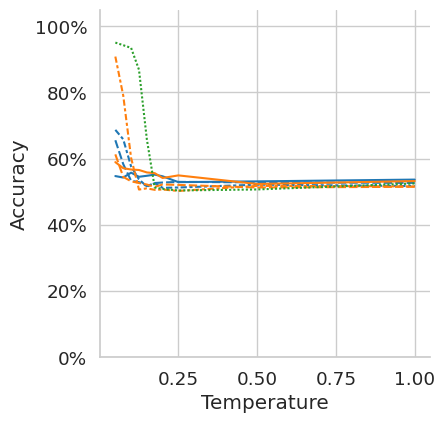}
         \caption{Test-ID Accuracy}
         \label{fig:test_id_projector_head}
     \end{subfigure}
     \hfill
     \begin{subfigure}[b]{0.22\textwidth}
         \centering
         \includegraphics[width=\textwidth]{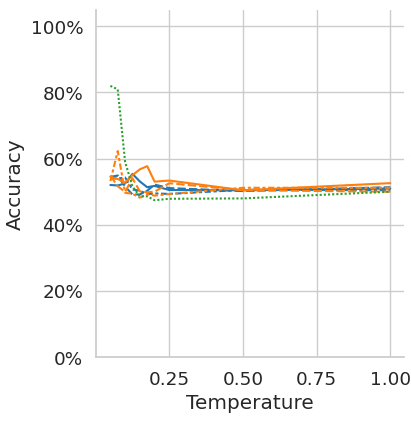}
         \caption{Test-OOD Accuracy}
         \label{fig:test_ood_projector_head}
     \end{subfigure}
     \hfill
     \begin{subfigure}[b]{0.445\textwidth}
         \centering
         \includegraphics[width=\textwidth]{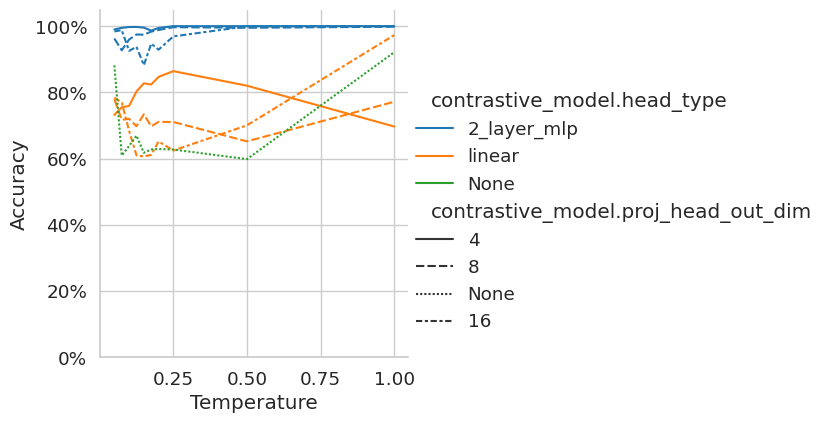}
         \caption{Domain Test-ID Accuracy}
         \label{fig:domain_test_id_projector_head}
     \end{subfigure}
        \caption{Projector head ablation of \textit{Standard Contrastive Learning} method. We consider 3 different types of projector head -- 2-layer MLP, linear, and no projector head; 3 different dimensions for the output of the projector head $\{4,8,16\}$; and, temperature $\tau \in \{0.05,0.075,0.1,0.125,0.15,0.175,0.2,0.25,0.5,1\}$. An encoder embedding dimension of 16-d, color variance of 25, and label fraction of 10\% are used for this experiment.}
        \label{fig:projector_head_ablation}
\end{figure}

In Figure~\ref{fig:projector_head_ablation} we see that introducing a projector head into the contrastive learning model on our MNIST dataset seems to deteriorate the performance of the \textit{Standard Contrastive Learning} method on both the in-distribution (Test-ID) and out-of-distribution (Test-OOD) domains. We also see that the accuracy of a domain discriminator trained on the learned embeddings (Domain Test-ID) seems to increase with projector head dimension (Figure~\ref{fig:domain_test_id_projector_head}). This suggests that the inclusion of the projector head leads to a less domain-invariant embedding space at the output layer of the encoder (pre-projection head). We hypothesize that this is because when a projection head is included, the model can use it to encode domain invariance instead of enforcing invariance through the encoder weights; thus, the invariance is lost when the projector head is discarded before downstream evaluation. Emerging insights on the role of the projector head in contrastive learning may further explain this observation \cite{balestriero2023cookbook}. We also note that this finding may not hold when moving to real-world datasets where the features mediated by domain are more complex than the color we use in this synthetic MNIST dataset.

\subsection{Effective Temperature Distributions -- Example Training Dynamics}
\label{sec:app_eff_temp}

\begin{figure}[t]
  \centering
  \includegraphics[width=0.9\linewidth]{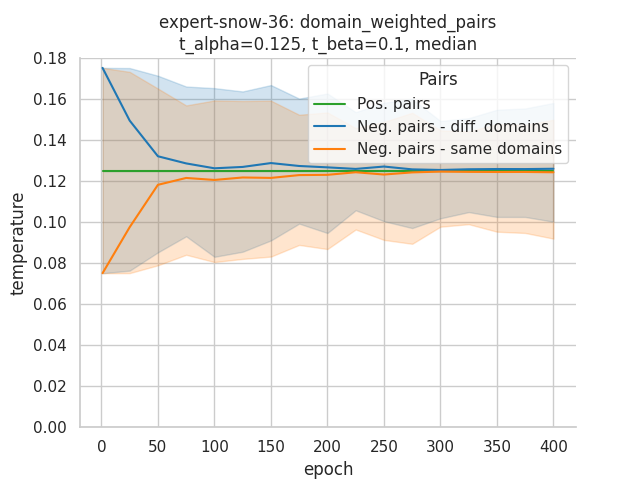}
  \caption{Distribution of \textit{effective temperatures} by pre-training epoch for a model with $\tau_\alpha = 0.125$ and $\tau_\beta = 0.1$ and the domain-weighted pairs approach. The thick lines represent the median values and the shading represents the range between 5th and 95th percentiles. We see that as pre-training progresses and the embedding space becomes more domain-invariant, the distribution of temperatures converges to the baseline temperature $\tau_\alpha$.}
  \label{fig:eff_temp_line}
\end{figure}

\begin{figure}[t]
  \centering
  \includegraphics[width=0.9\linewidth]{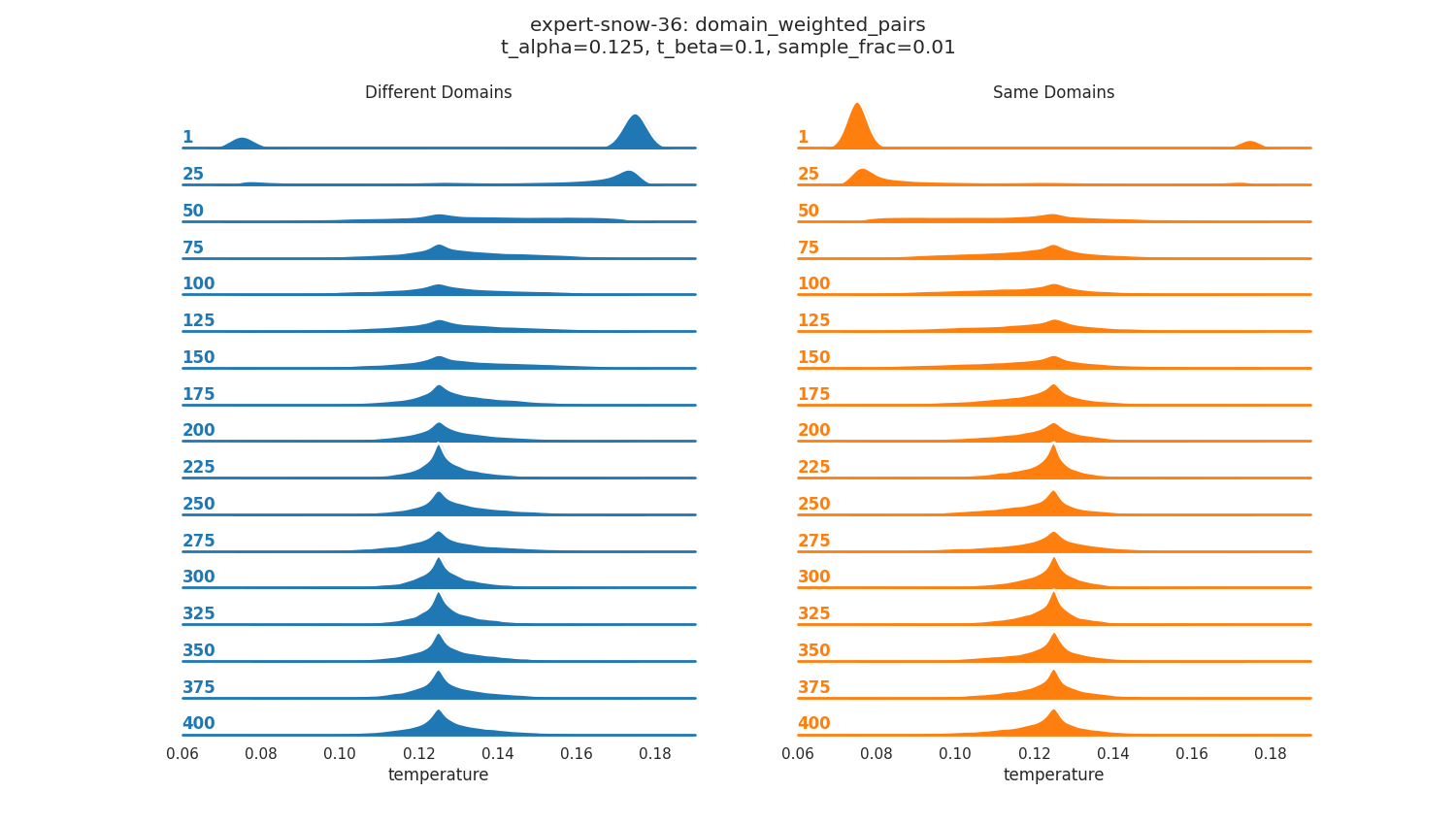}
  \caption{Distribution of \textit{effective temperatures} by pre-training epoch for a model with $\tau_\alpha = 0.125$ and $\tau_\beta = 0.1$ and the domain-weighted pairs approach. We see that as pre-training progresses and the embedding space becomes more domain-invariant, the distribution of temperatures converges to the baseline temperature $\tau_\alpha$. The pre-training epoch are the numbers to the left-hand side of each axis.}
  \label{fig:eff_temp_ridge}
\end{figure}

Figures~\ref{fig:eff_temp_line} \& ~\ref{fig:eff_temp_ridge} show the distribution of \textit{effective temperatures} by pre-training epoch. These confirm empirically that the proposed loss function is working as expected -- i.e., it initially uses a lower temperature on pairs of samples from more similar domains, but over time its distribution converges to the baseline temperature $\tau_\alpha$ as domain information is removed from the embedding space.

\end{document}